\documentclass[11pt,a4paper]{article}
\usepackage[dvipsnames]{xcolor}
\usepackage[hyperref]{acl2021}
\usepackage{textcomp}
\usepackage{times}
\usepackage{latexsym}

\usepackage{microtype}

\usepackage{amsmath, bm, amsfonts}
\usepackage{graphicx, subfigure}
\usepackage{booktabs, multirow}
\usepackage{array}

\aclfinalcopy 
\def\aclpaperid{580} 

\title{Syntax-Enhanced Pre-trained Model}

\author{Zenan Xu$^{1}$\thanks{\ \ \ Work is done during internship at Microsoft.}$^{~~\dag}$, Daya Guo$^{1*}$, Duyu Tang$^{2}$\thanks{~~~For questions,  please contact D. Tang and Z. Xu.}~, Qinliang Su$^{1,4,5}$\thanks{~~~Corresponding author.}~, Linjun Shou$^{3}$,\\
	\bf{Ming Gong$^{3}$, Wanjun Zhong$^{1*}$, Xiaojun Quan$^{1}$, Daxin Jiang$^{3}$, and  Nan Duan$^{2}$}  \\
	$^1$School of Computer Science and Engineering, Sun Yat-sen University, Guangzhou, China \\
	$^2$Microsoft Research Asia, Beijing, China\\
	$^3$Microsoft Search Technology Center Asia, Beijing, China\\
	$^4$Guangdong Key Laboratory of Big Data Analysis and Processing, Guangzhou, China\\
	$^5$Key Lab. of Machine Intelligence and Advanced Computing, Ministry of Education, China\\
	{\tt\{xuzn, guody5, zhongwj25\}@mail2.sysu.edu.cn} \\ 
	{\tt\{suqliang, quanxj3\}@mail.sysu.edu.cn} \\
	{\tt\{dutang,lisho,migon,djiang,nanduan\}@microsoft.com}}
\date{}

\begin{document}
\maketitle
\begin{abstract}
	We study the problem of leveraging the syntactic structure of text to enhance pre-trained models such as BERT and RoBERTa. Existing methods utilize syntax of text either in the pre-training stage or in the fine-tuning stage, so that they suffer from discrepancy between the two stages.
	Such a problem would lead to the necessity of having human-annotated syntactic information, which limits the application of existing methods to broader scenarios.
	To address this, we present a model that utilizes the syntax of text in both pre-training and fine-tuning stages.      
	Our model is based on Transformer with a syntax-aware attention layer that considers the dependency tree of the text.
	We further introduce a new pre-training task of predicting the syntactic distance among tokens in the dependency tree. 
	We evaluate the model on three downstream tasks, including relation classification, entity typing, and question answering. Results show that our model achieves state-of-the-art performance on six public benchmark datasets. 
	We have two major findings.
	First, we demonstrate that infusing automatically produced syntax of text improves pre-trained models.
	Second, \textit{global} syntactic distances among tokens bring larger performance gains compared to \textit{local} head relations between contiguous tokens.\footnote{The source data is available at \href{https://github.com/Hi-ZenanXu/Syntax-Enhanced_Pre-trained_Model}{https://github.com/Hi-ZenanXu/Syntax-Enhanced\_Pre-trained\_Model.}}
\end{abstract}

\section{Introduction}
	Pre-trained models such as BERT \cite{devlin2019bert}, GPT \cite{radford2018improving}, and RoBERTa \cite{liu2019roberta} have advanced the state-of-the-art performances of various natural language processing tasks. The successful recipe is that a model is first pre-trained on a huge volume of unsupervised data with self-supervised objectives, and then is fine-tuned on supervised data with the same data scheme. Dominant pre-trained models represent a text as a sequence of tokens\footnote{Such tokens can be words or word pieces. We use token for clarity.}. The merits are that such basic text representations are available from vast amounts of unsupervised data, and that models pre-trained and fine-tuned with the same paradigm usually achieve good accuracy in practice \cite{guu2020realm}. However, an evident limitation of these methods is that richer syntactic structure of text is ignored.
	
	In this paper, we seek to enhance pre-trained models with syntax of text. Related studies attempt to inject syntax information either only in the fine-tuning stage \cite{nguyen2020tree, sachan2020syntax}, or only in the pre-training stage \cite{wang2020k}, which results in discrepancies. When only fusing syntax information in the fine-tuning phase, \citet{sachan2020syntax} finds that there is no performance boost unless high quality human-annotated dependency parses are available. However, this requirement would limit the application of the model to broader scenarios where human-annotated dependency information is not available.
	
	To address this, we conduct a large-scale study on injecting automatically produced syntax of text in both the pre-training and fine-tuning stages. We construct a pre-training dataset by applying an off-the-shelf dependency parser \cite{qi2020stanza} to one billion sentences from common crawl news. With these data, we introduce a syntax-aware pre-training task, called dependency distance prediction, which predicts the syntactic distance between tokens in the dependency structure. Compared with the pre-training task of dependency head prediction \cite{wang2020k} that only captures local syntactic relations among words, dependency distance prediction leverages global syntax of the text. In addition, we developed a syntax-aware attention layer, which can be conveniently integrated into Transformer \cite{vaswani2017attention} to allow tokens to selectively attend to contextual tokens based on their syntactic distance in the dependency structure. 
	
	We conduct experiments on entity typing, question answering and relation classification on six benchmark datasets. Experimental results show that our method achieves state-of-the-art performance on all six datasets. Further analysis shows that our model can indicate the importance of syntactic information on downstream tasks, and that the newly introduced dependency distance prediction task could capture the global syntax of the text, performs better than dependency head prediction. In addition, compared with experimental results of injecting syntax information in either the pre-training or fine-tuning stage, injecting syntax information in both stages achieves the best performance.
	
	In summary, the contribution of this paper is threefold. (1) We demonstrate that infusing automatically produced dependency structures into the pre-trained model shows superior performance over downstream tasks. (2) We propose a syntax-aware attention layer and a pre-training task for infusing syntactic information into the pre-trained model. (3) We find that the newly introduced dependency distance prediction task performs better than the dependency head prediction task.

	\begin{figure*}[!tp]
		\centering
		\includegraphics[scale=0.6]{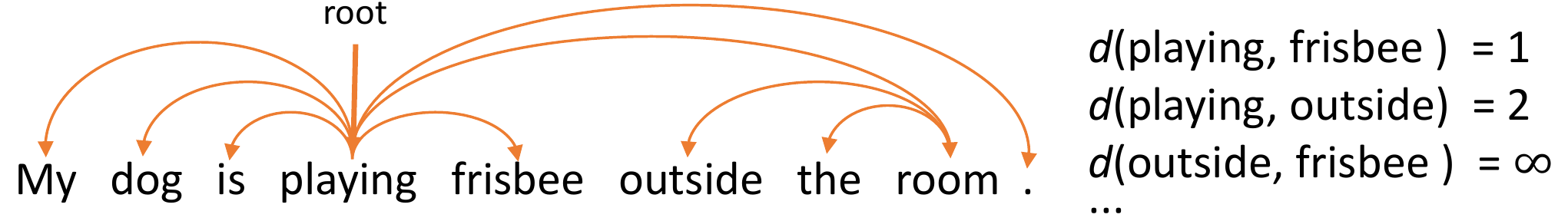}
		\caption{The dependency tree of the sentence, ``My dog is playing frisbee outside the room," after running the Stanza parser.} 
		\label{fig:distance}
		\vspace{-2mm}
	\end{figure*}
	
	\section{Related Work}
	Our work involves injecting syntax information into pre-trained models. First, we will review recent studies on analyzing the knowledge presented in pre-trained models, and then we will introduce the existing methods that enhance pre-trained models with syntax information.
	
	\subsection{Probing Pre-trained Models}
	With the huge success of pre-trained models \cite{devlin2019bert, radford2018improving} in a wide range of NLP tasks, lots of works study to what extent pre-trained models inherently. Here, we will introduce recent works on probing linguistic information, factual knowledge, and symbolic reasoning ability from pre-trained models respectively. In terms of linguistic information, \citet{syntax_probe_bert_hewitt2019} learn a linear transformation to predict the depth of each word on a syntax tree based on their representation, which indicates that the syntax information is implicitly embedded in the BERT model. However, \citet{treetransformer_treeintoselfatt} find that the attention scores calculated by pre-trained models seem to be inconsistent with human intuitions of hierarchical structures, and indicate that certain complex syntax information may not be naturally embedded in BERT. In terms of probing factual knowledge, \citet{Petroni2019LanguageMA} find that pre-trained models are able to answer fact-filling cloze tests, which indicates that the pre-trained models have memorized factual knowledge. However, \citet{Poerner2019BERTIN} argue that BERT's outstanding performance of answering fact-filling cloze tests is partly due to the reasoning of the surface form of name patterns. In terms of symbolic reasoning, \citet{talmor2020olmpics} test the pre-trained models on eight reasoning tasks and find that the models completely fail on half of the tasks. Although probing knowledge from pre-trained model is a worthwhile area, it runs perpendicular to infusing knowledge into pre-trained models.
	
	\subsection{Integrating Syntax into Pre-trained Models}
	Recently, there has been growing interest in enhancing pre-trained models with syntax of text. Existing methods attempt to inject syntax information in the fine-tuning stage or only in the pre-training stage. We first introduce related works that inject syntax in the fine-tuning stage. \citet{nguyen2020tree} incorporate a tree-structured attention into the Transformer framework to help encode syntax information in the fine-tuning stage. \citet{zhang2020sg} utilize the syntax to guide the Transformer model to pay no attention to the dispensable words in the fine-tuning stage and improve the performance in machine reading comprehension. \citet{sachan2020syntax} investigate two distinct strategies for incorporating dependency structures in the fine-tuning stage and obtain state-of-the-art results on the semantic role labeling task. Meanwhile, \citet{sachan2020syntax} argue that the performance boost is mainly contributed to the high-quality human-annotated syntax. However, human annotation is costly and difficult to extend to a wide range of applications. Syntax information can also be injected in the pre-training stage. \citet{wang2020k} introduce head prediction tasks to inject syntax information into the pre-trained model, while syntax information is not provided during inference. Note that the head prediction task in \citet{wang2020k} only focuses on the local relationship between two related tokens, which prevents each token from being able to perceive the information of the entire tree. Despite the success of utilizing syntax information, existing
	methods only consider the syntactic information
	of text in the pre-training or the fine-tuning
	stage so that they suffer from discrepancy between the pre-training and the fine-tuning stage. To bridge this gap, we conduct a large-scale study on injecting automatically produced syntax information in both the two stages. Compared with the head prediction task \cite{wang2020k} that captures the local relationship, we introduce the dependency distance prediction task that leverages the global relationship to predict the distance of two given tokens.
	
	\section{Data Construction}
	In this paper, we adopt the dependency tree to express the syntax information. Such a tree structure is concise and only expresses necessary information for the parse \cite{jurafsky2000speech}. Meanwhile, its head-dependent relation can be viewed as an approximation to the semantic relationship between tokens, which is directly useful for capturing semantic information. The above advantages help our model make more effective use of syntax information. 
	Another available type of syntax information is the constituency tree, which is used in \citet{nguyen2020tree}. However, as pointed out in \citet{jurafsky2000speech}, the relationships between the tokens in dependency tree can directly reflect important syntax information, which is often buried in the more complex constituency trees. This property requires extra techniques to extracting relation among the words from a constituency tree \citep{jurafsky2000speech}\footnote{\href{https://web.stanford.edu/~jurafsky/slp3/}{https://web.stanford.edu/\~{}jurafsky/slp3/}}. 
	
	The dependency tree takes linguistic words as one of its basic units. However, most pre-trained models take subwords (also known as the word pieces) instead of the entire linguistic words as the input unit, and this necessitates us to extend the definition of the dependency tree to include subwords. Following \citet{wang2020k}, we will add edges from the first subword of $v$ to all subwords of $u$, if there exists a relationship between linguistic word $v$ and word $u$.
	
	Based on the above extended definition, we build a pre-training dataset from open-domain sources. Specifically, we randomly collect 1B sentences from publicly released common crawl news datasets \cite{NIPS2019_9106} that contain English news articles crawled between December 2016 and March 2019. Considering its effectiveness and ability to expand to multiple languages, we adopt off-the-shelf Stanza\footnote{\href{https://github.com/stanfordnlp/stanza}{https://github.com/stanfordnlp/stanza}} to automatically generate the syntax information for each sentence. The average token length of each sentence is 25.34, and the average depth of syntax trees is 5.15.
	
	\section{Methodology}
	In this section, we present the proposed \textbf{S}yntax-\textbf{E}nhanced \textbf{PRE}-trained \textbf{M}odel (\textbf{SEPREM}). We first define the syntax distance between two tokens. Based on the syntax distance, we then introduce a syntax-aware attention layer to learn syntax-aware representation and a pre-training task to enable model to capture global syntactic relations among tokens.
	
	\subsection{Syntax Distance over Syntactic Tree} 
	Intuitively, the distance between two tokens on the syntactic tree may reflect the strength of their linguistic correlation. If two tokens are far away from each other on the syntactic tree, the strength of their linguistic correlation is likely weak. Thus, we define the distance of two tokens over the dependency tree as their syntactic distance. Specifically, we define the distance between the token $v$ and token $u$ as 1, i.e. $d(v, u)=1$, if $v$ is the head of $u$. If two tokens are not directly connected in the dependency graph, their distance is the summation of the distances between adjacent nodes on the path. If two tokens are separated in the graph, their distance is set to infinite. Taking the sentence ``\textit{My dog is playing frisbee outside the room.}" in Fig \ref{fig:distance} as an example, $d(\textit{playing}, \textit{frisbee})$ equals 1 since the token ``\textit{playing}" is the head of the token ``\textit{frisbee}".

	\subsection{Syntax-Aware Transformer}
	\label{Syntax-Aware Transformer}
	We follow BERT \cite{devlin2019bert} and use the multi-layer bidirectional Transformer \cite{vaswani2017attention} as the model backbone. The model takes a sequence $X$ as the input and applies $N$ transformer layers to produce contextual representation:
	\begin{equation} \label{equ:transformer}
	\bm{H}^n=transformer_n((1-\alpha)\bm{H}^{n-1}+\alpha\bm{\hat{H}}^{n-1})
	\end{equation}
	where $n \in [1,N]$ denotes the $n$-th layer of the model, $\bm{\hat{H}}$ is the syntax-aware representation which will be described in Section \ref{sec:distance-aware attention layer}, $\bm{H}^0$ is embeddings of the sequence input $X$, and $\alpha$ is a learnable variable. 
	
	However, the introduction of syntax-aware representation $\bm{\hat{H}}$ in the Equation \ref{equ:transformer} changes the architecture of Transformer, invalidating the original weights from pre-trained model, such as BERT and RoBERTa. Instead, we introduce a learnable importance score $\alpha$ that controls the proportion of integration between contextual and syntax-aware representation. When $\alpha$ is equal to zero, the syntax-aware representation is totally excluded and the model is architectural identical to vanilla Transformer. Therefore, we initialize the parameter $\alpha$ as the small but not zero value, which can help better fuse syntactic information into existing pre-trained models. We will discuss importance score $\alpha$ in detailed in Section \ref{model analysis}. 
	
	Each transformer layer $transformer_n$ contains an architecturally identical transformer block, which is composed of a multi-headed self-attention $MultiAttn$ \cite{vaswani2017attention} and a followed feed forward layer $FFN$. Formally, the output $\bm{H}^n$ of the transformer block $transformer_n(H'_{n-1})$ is computed as:
	\begin{equation}\label{equa:transformer-block}
	\begin{split}
	&\bm{G}'_n=LN(MultiAttn(H'_{n-1})+H'_{n-1})\\
	&\bm{H}^n=LN(FFN(\bm{G}'_n)+\bm{G}'_n)\\
	\end{split}
	\end{equation}   
	where  the input $H'_{n-1}$ is $(1-\alpha) \bm{H}^{n-1} + \alpha\bm{\hat{H}}^{n-1}$ and $LN$ represents a layer normalization operation.
	
	\subsection{Syntax-aware Attention Layer}
	\label{sec:distance-aware attention layer}
	In this section, we will introduce how to obtain the syntax-aware representation $\bm{\hat{H}}$ used in syntax-aware transformer.
	
	\paragraph{Tree Structure Encoding}
	We adopt a distance matrix \bm{$D$} to encode the tree structure. The advantages of distance matrix \bm{$D$} are that it can well preserve the hierarchical syntactic structure of text and can directly reflect the distance of two given tokens.  Meanwhile, its uniqueness property guarantees the one-to-one mapping of the tree structure. Given a dependency tree, the element $\bm{D}_{i,j}$ of distance matrix \bm{$D$} in $i$-th row and $j$-th column is defined as:
	\begin{equation}
	\bm{D}_{i,j} = \left\{
	\begin{array}{cl}
	d(i,j), & \text{if exists a path from $v_i$ to $v_j$}, \\
	0, & \text{if $i$ = $j$ and otherwise}.
	\end{array}
	\right.
	\end{equation}
	where $v_i$ and $v_j$ are tokens on the dependency tree.
	Based on the concept that distance is inversely proportional to importance, we normalize the matrix \bm{$D$} and obtain the normalized correlation strength matrix \bm{$\tilde{D}$} as follows:
	\begin{equation}
	\bm{\tilde{D}}_{i,j} = \left\{
	\begin{array}{cl}
	\frac{1 / \bm{D}_{i,j}}{\sum_{z\in\{y|\bm{D}_{i,y} \neq 0\}} (1/\bm{D}_{i,z})}, & \text{if $\bm{D}_{i,j} \neq 0$}, \\
	0, & \text{otherwise}.
	\end{array}
	\right.
	\end{equation}
	
	\paragraph{Syntax-aware Representation}
	Given the tree structure representation $\bm{\tilde{D}}$ and the contextual representation $\bm{H}^n$, we fuse the tree structure into the contextual representation as:
	\begin{equation}
	\label{diatance-aware representation}
	\bm{\hat{H}}^n = \sigma(\bm{W}_n^1\bm{H}^n + \bm{W}_n^2\bm{\tilde{D}}\bm{H}^n)
	\end{equation}
	where $\sigma$ is the activation function, $\bm{W}_n^1$ and $\bm{W}_n^2\in\mathbb{R}^{d_h \times d_h}$ are model parameters. We can see that $\bm{\tilde{D}}\bm{H}^n$ allows one to aggregate information from others along the tree structure. The closer they are on the dependency tree, the larger the attention weight, and thus more information will be propagated to each other, and vice verse.
	
	\subsection{Syntax-aware Pre-training Task}
	\label{sec:Syntax-aware_Pre-training_Tasks}
	To better understand the sentences, it is beneficial for model to be aware of the underlying syntax. To this end, a new pre-training task, named dependency distance prediction task (DP), is designed to enhance the model's ability of capturing global syntactic relations among tokens. Specifically, we first randomly mask some elements in the distance matrix $\bm{D}$, e.g., supposed $\bm{D}_{i,j}$. Afterwards, the representations of tokens $i\text{ and }j$ from SEPREM are concatenated and fed into a linear classifier, which outputs the probabilities over difference distances. In all of our experiments, 15\% of distance are masked at random.
	
	Similar to BERT \cite{devlin2019bert} and RoBERTa \cite{liu2019roberta}, we conduct the following operations to boost the robustness. The distance in matrix $\bm{D}$ will be masked at 80\% probability or replaced by a random integer with a probability of 10\%. For the rest 10\% probability, the distance will be maintained. 
	
	During pre-training, in addition to the DP pre-training task, we also use the dependency head prediction (HP) task, which is used in \citet{wang2020k} to capture the local head relation among words, and the dynamic masked language model (MLM), which is used in \citet{liu2019roberta} to capture contextual information. The final loss for the pre-training is the summation of the training loss of DP, HP and MLM tasks.
	
	\subsection{Implementation Details}
	The implementation of SEPREM is based on HuggingFace's Transformer \citep{wolf2019huggingface}. To accelerate the training process, we initialize parameters from RoBERTa model released by HuggingFace\footnote{\href{https://huggingface.co/transformers/}{https://huggingface.co/transformers/}}, which contains 24 layers, with 1024 hidden states in each layer. The number of parameters of our model is 464M. We pre-train our model with 16 32G NVIDIA V100 GPUs for approximately two weeks. The batch size is set to 2048, and the total steps are 500000, of which 30000 is the warm up steps.
	
	In both pre-training and fine-tuning stages, our model takes the syntax of the text as the additional input, which is pre-processed in advance. Specially, we obtain the dependency tree of each sentence via Stanza and then generate the normalized distance matrix.
	
	\section{Experiments}
	In this section, we evaluate the proposed SEPREM on six benchmark datasets over three downstream tasks, {\it i.e.}, entity typing, question answering and relation classification.
	
	\subsection{Entity Typing}
	The entity typing task requires the model to predict the type of a given entity based on its context. Two fine-grained public datasets, Open Entity \citep{choi2018openentity} and FIGER \citep{ling2015figer}, are employed to evaluate our model. The statistics of the aforementioned datasets are shown in Table \ref{table:statistic_oe_rc}. Following \citet{wang2020k}, special token ``@" is added before and after a certain entity, then the representation of the first special token ``@" is adopted to predict the type of the given entity. To keep the evaluation criteria consistent with previous works \citep{shimaoka2016attentive, zhang2019ernie, peters2019knowledge, wang2019kepler, xiong2019pretrained}, we adopt loose micro precision, recall, and F1 to evaluate model performance on Open Entity datasets. As for FIGER datasets, we utilize strict accuracy, loose macro-F1, and loose micro-F1 as evaluation metrics.
	
	\begin{table}[!t]
		\small
		\begin{center}
			\begin{tabular}{@{}l|cccc@{}}
				\toprule
				Dataset     & Train     & Dev    & Test   & Label\\ \midrule
				Open Entity & 2,000     & 2,000  & 2,000  & 6             \\
				FIGER       & 2,000,000 & 10,000 & 563    & 113           \\ 
				\midrule
				TACRED      & 68,124    & 22,631 & 15,509 & 42            \\ 
				\bottomrule
			\end{tabular}
		\end{center}
		\caption{The statistics of the entity typing datasets, i.e., Open Entity and FIGER, and relation classification dataset TACRED. Label refers to type of a given entity or relation between two entities.}
		\label{table:statistic_oe_rc}
		\vskip -0.15in
	\end{table}
	
	\paragraph{Baselines} 
	NFGEC \citep{shimaoka2016attentive} recursively composes representation of entity context and further incorporates an attention mechanism to capture fine-grained category memberships of an entity. KEPLER \citep{wang2019kepler} infuses knowledge into the pre-trained models and jointly learns the knowledge embeddings and language representation. RoBERTa-large (continue training) learns on the proposed pre-training dataset under the same settings with SEPREM but only with dynamic MLM task. In addition, we also report the results of BERT-base \citep{devlin2019bert}, ERNIE \citep{zhang2019ernie}, KnowBERT \citep{peters2019knowledge}, WKLM \citep{xiong2019pretrained}, RoBERTa-large, and K-adapter \citep{wang2020k} for a full comparison.

	\begin{table*}[!t]
		\centering
		\begin{tabular}{@{}l|ccc|ccc@{}}
			\toprule
			\multirow{2}{*}{\textbf{Model}} & \multicolumn{3}{c|}{\textbf{OpenEntity}} & \multicolumn{3}{c}{\textbf{FIGER}} \\ \cmidrule(l){2-7} 
			& \textbf{P} & \textbf{R} & \textbf{Mi-$\textbf{F}_1$} & \textbf{Acc} & \textbf{Ma-$\textbf{F}_1$} & \textbf{Mi-$\textbf{F}_1$} \\ \midrule
			NFGEC \cite{shimaoka2016attentive} & 68.80 & 53.30 & 60.10 & 55.60 & 75.15 & 71.73 \\ 
			BERT-base \cite{zhang2019ernie} & 76.37 & 70.96 & 73.56 & 52.04 & 75.16 & 71.63 \\
			ERNIE \cite{zhang2019ernie} & 78.42 & 72.90 & 75.56 & 57.19 & 75.61 & 73.39 \\
			KnowBERT \cite{peters2019knowledge} & 78.60 & 73.70 & 76.10 & - & - & - \\
			KEPLER \cite{wang2019kepler} & 77.20 & 74.20 & 75.70 & - & - & - \\
			WKLM \cite{xiong2019pretrained} & - & - & - & 60.21 & 81.99 & 77.00 \\
			K-Adapter \cite{wang2020k} & 79.25 & 75.00 & 77.06 & 61.81 & 84.87 & 80.54 \\
			\midrule
			RoBERTa-large & 77.55 & 74.95 & 76.23 & 56.31 & 82.43 & 77.83 \\
			RoBERTa-large (continue training) & 77.63 & 75.01 & 76.30 & 56.52 & 82.37 & 77.81\\
			
			SEPREM & \textbf{81.07} & \textbf{77.14} & \textbf{79.06} & \textbf{63.21} & \textbf{86.14} & \textbf{82.05}\\
			\bottomrule
		\end{tabular}%
		\vspace{-2mm}
		\caption{Results for entity typing task on the OpenEntity and FIGER datasets.}
		\label{tab:entity_typing}
	\end{table*}
	
	\paragraph{Experimental Results} As we can see in Table \ref{tab:entity_typing}, our SEPREM outperforms all other baselines on both entity typing datasets. In the Open Entity dataset, with the utility of the syntax of text, SEPREM achieves an improvement of 3.6\% in micro-F1 score comparing with RoBERTa-large (continue training) model. The result demonstrates that the proposed syntax-aware pre-training tasks and syntax-aware attention layer help to capture the syntax of text, which is beneficial to predict the types more accurately. As for the FIGER dataset, which contains more labels about the type of entity, SEPREM still brings an improvement in strict accuracy, macro-F1, and micro-F1. This demonstrates the effectiveness of leveraging syntactic information in tasks with more fine-grained information. Specifically, compared with the K-adapter model, our SEPREM model brings an improvement of 2.6\% F1 score on Open Entity dataset. It is worth noting that SEPREM model is complementary to the K-adapter model, both of which inject syntactic information into model during pre-training stage. This improvement indicates that injecting syntactic information in both the pre-training and fine-tuning stages can make full use of the syntax of the text, thereby benefiting downstream tasks.
	
	\begin{table}[!t]
		\begin{center}
			\begin{tabular}{@{}l|ccc@{}}
				\toprule
				Dataset     & Train     & Dev    & Test \\ 
				\midrule
				SearchQA & 99,811     & 13,893  & 27,247 \\
				Quasar-T & 28,496 & 3,000 & 3,000        \\
				\midrule
				CosmosQA & 25,588     &3,000  & 7,000   \\
				\bottomrule
			\end{tabular}%
		\end{center}
		\caption{The statistics of the question answering datasets: SearchQA, Quasar-T and CosmosQA.}
		\label{table:statistic_qa}
		\vskip -0.15in
	\end{table}
	
	\subsection{Question Answering}
	We use open-domain question answering (QA) task and commonsense QA task to evaluate the proposed model. Open-domain QA requires models to answer open-domain questions with the help of external resources such as materials of collected documents and webpages. We use SearchQA \citep{dunn2017searchqa} and QuasarT \citep{dhingra2017quasar} for this task, and adopt ExactMatch (EM) and loose F1 scores as evaluation metrics. In this task, we first retrieve related paragraphs according to the question from external materials via the information retrieval system, and then a reading comprehension technique is adopted to extract possible answers from the above retrieved paragraphs. Following previous work \citep{lin2018denoising}, we use the retrieved paragraphs provided by \citet{wang2017gated} for the two datasets. 
	For fair comparison, we follow \citet{wang2020k} to use \resizebox{\hsize}{!}{$[$$<$$sep$$>$$, quesiton, $$<$$/sep$$>$$,paragraph,$$<$$/sep$$>$$]$}
	as the input, where $<$$sep$$>$ is a special token in front of two segmants and $<$$/sep$$>$ is a special symbol to split two kinds of data types. We take the task as a multi-classification to fine-tune the model and use two linear layers over the last hidden features from models to predict the start and end positions of the answer span.
	
	Commonsense QA aims to answer questions which require commonsense knowledge that is not explicitly expressed in the question. 
	We use the public CosmosQA dataset \citep{huang2019cosmos} for this task, and the accuracy scores are used as evaluation metrics. The data statistics of the above three datasets are shown in Table \ref{table:statistic_qa}. In CosmosQA, each question has 4 candidate answers, and we concatenate the question together with each answer separately as $[$$<$$sep$$>$$, context, $$<$$/sep$$>$$,paragraph,$$<$$/sep$$>$$]$ for input. The representation of the first token is adopted to calculate a score for this answer, and the answer with the highest score is regarded as the prediction answer for this question.
	
	\begin{table*}[!t]
		\centering
		\small
		\begin{tabular}{@{}l|cc|cc|c@{}}
			\toprule
			\multirow{2}{*}{\textbf{Model}} & \multicolumn{2}{c|}{\textbf{SearchQA}} & \multicolumn{2}{c|}{\textbf{Quasar-T}} & \textbf{CosmosQA} \\ \cmidrule(l){2-6} & \textbf{EM} & \textbf{$\textbf{F}_1$} & \textbf{EM} & \textbf{$\textbf{F}_1$} & \textbf{Accuracy} \\ 
			\midrule
			BiDAF \citep{SeoKFH2017} & 28.60 & 34.60 & 25.90 & 28.50 & -\\
			AQA \citep{Buck2017AskTR} & 40.50  & 47.40 & - & -  & -\\
			R\textasciicircum{}3 \citep{wang2017reinforced} & 49.00 & 55.30 & 35.30 & 41.70 & -\\
			DSQA \citep{lin2018denoising} & 49.00 & 55.30 & 42.30 & 49.30 & -\\
			Evidence Agg. \citep{wang2017evidence} & 57.00 & 63.20 & 42.30 & 49.60 &  -\\
			BERT \citep{xiong2019pretrained} & 57.10 & 61.90 & 40.40 & 46.10 & -\\
			WKLM \citep{xiong2019pretrained} & 58.70 & 63.30 & 43.70 & 49.90 & -\\
			WKLM + Ranking \citep{xiong2019pretrained} & 61.70 & 66.70  & 45.80 & 52.20 & -\\
			$\text{BERT-FT}_{RACE+SWAG}$ \citep{huang2019cosmos} & - & - & - & - & 68.70\\ 
			\textsc{K-Adapter} \citep{wang2020k} & 61.96 & 67.31 & 45.69 & 52.48 & 81.83\\
			\midrule
			RoBERTa-large & 59.01 & 65.62 & 40.83 & 48.84 & 80.59\\
			RoBERTa-large (continue training) & 59.34 & 65.71 & 40.91 & 49.04 & 80.75\\
			SEPREM & \textbf{62.31} & \textbf{67.74} & \textbf{46.37} & \textbf{53.18} & \textbf{82.37}\\
			\bottomrule
		\end{tabular}
		\caption{Results on QA datasets including: SearchQA, Quasar-T and CosmosQA.}
		\label{tab:question_answering}
		\vskip -0.15in
	\end{table*}
	
	\begin{table}[!t]
		\centering
		\resizebox{\linewidth}{!}{
			\begin{tabular}{@{}lccc@{}}
				\toprule
				\textbf{Model} & \textbf{P} & \textbf{R} & \textbf{$\textbf{F}_1$} \\ \midrule
				C-GCN \citep{zhang2018graph} & 69.90 & 63.30 & 66.40 \\
				BERT-base \citep{zhang2019ernie} & 67.23 & 64.81 & 66.00 \\
				ERNIE \citep{zhang2019ernie} & 69.97 & 66.08 & 67.97 \\
				BERT-large \citep{soares2019matching} & - & - & 70.10 \\
				BERT+MTB \citep{soares2019matching} & - & - & 71.50 \\
				KnowBERT \citep{peters2019knowledge} & 71.60 & 71.40 & 71.50 \\
				KEPLER \citep{wang2019kepler} & 70.43 & 73.02 & 71.70 \\ 
				K-Adapter \citep{wang2020k} & 70.05 & 73.92 & 71.93\\ 
				\midrule
				RoBERTa-large & 70.17 & 72.36 & 71.25 \\
				RoBERTa-large (continue training) & 70.19 & 72.41 & 71.28\\
				SEPREM & \textbf{70.57} & \textbf{74.36} & \textbf{72.42}\\
				\bottomrule
		\end{tabular}}
		\caption{Results for relation classification task on TACRED dataset.}
		\label{table:RC}
		\vskip -0.2in
	\end{table}

	\paragraph{Baselines}
	BiDAF \citep{SeoKFH2017} is a bidirectional attention network to obtain query-aware context representation. AQA \citep{Buck2017AskTR} adopts a reinforce-guide questions re-write system and generates answers according to the re-written questions. R\textasciicircum{}3 \citep{wang2017reinforced} selects the most confident paragraph with a designed reinforcement ranker. DSQA \citep{lin2018denoising} employs a paragraph selector to remove paragraphs with noise and a paragraph reader to extract the correct answer from denoised paragraphs. Evidence Agg. \citep{wang2017evidence} makes use of multiple passages to generate answers. $\text{BERT-FT}_{RACE+SWAG}$ \citep{huang2019cosmos} sequentially fine-tunes the BERT model on the RACE and SWAG datasets for knowledge transfer. Besides the aforementioned models, we also report the results of BERT \citep{xiong2019pretrained}, WKLM \citep{xiong2019pretrained}, WKLM + Ranking \citep{xiong2019pretrained}, RoBERTa-large, RoBERTa-large (continue training), and K-Adapter \citep{wang2020k} for a detailed comparison.
	
	\paragraph{Experimental Results}
	The results of the open-domain QA task are shown in Table \ref{tab:question_answering}. We can see that the proposed SEPREM model brings significant gains of 3.1\% and 8.4\% in F1 scores, compared with RoBERTa-large (continue training) model. This may be partially attributed to the fact that, QA task requires a model to have reading comprehension ability \citep{wang2020k}, and the introduced syntax information can guide the model to avoid concentrating on certain dispensable words and improve its reading comprehension capacity \citep{zhang2020sg}. Meanwhile, SEPREM achieves state-of-the-art results on the CosmosQA dataset, which demonstrates the effectiveness of the proposed SEPREM model. It can be also seen that the performance gains observed in CosmosQA are not as substantial as those in the open-domain QA tasks. We speculate that CosmosQA requires capacity for contextual commonsense reasoning and the lack of explicitly injection of commonsense knowledge into SEPREM model limits its improvement. 
	
	\begin{figure*}[ht]
		\centering
		\subfigure[Open Entity]{
			\begin{minipage}[t]{0.33\linewidth}
				\centering
				\includegraphics[width=2.03in]{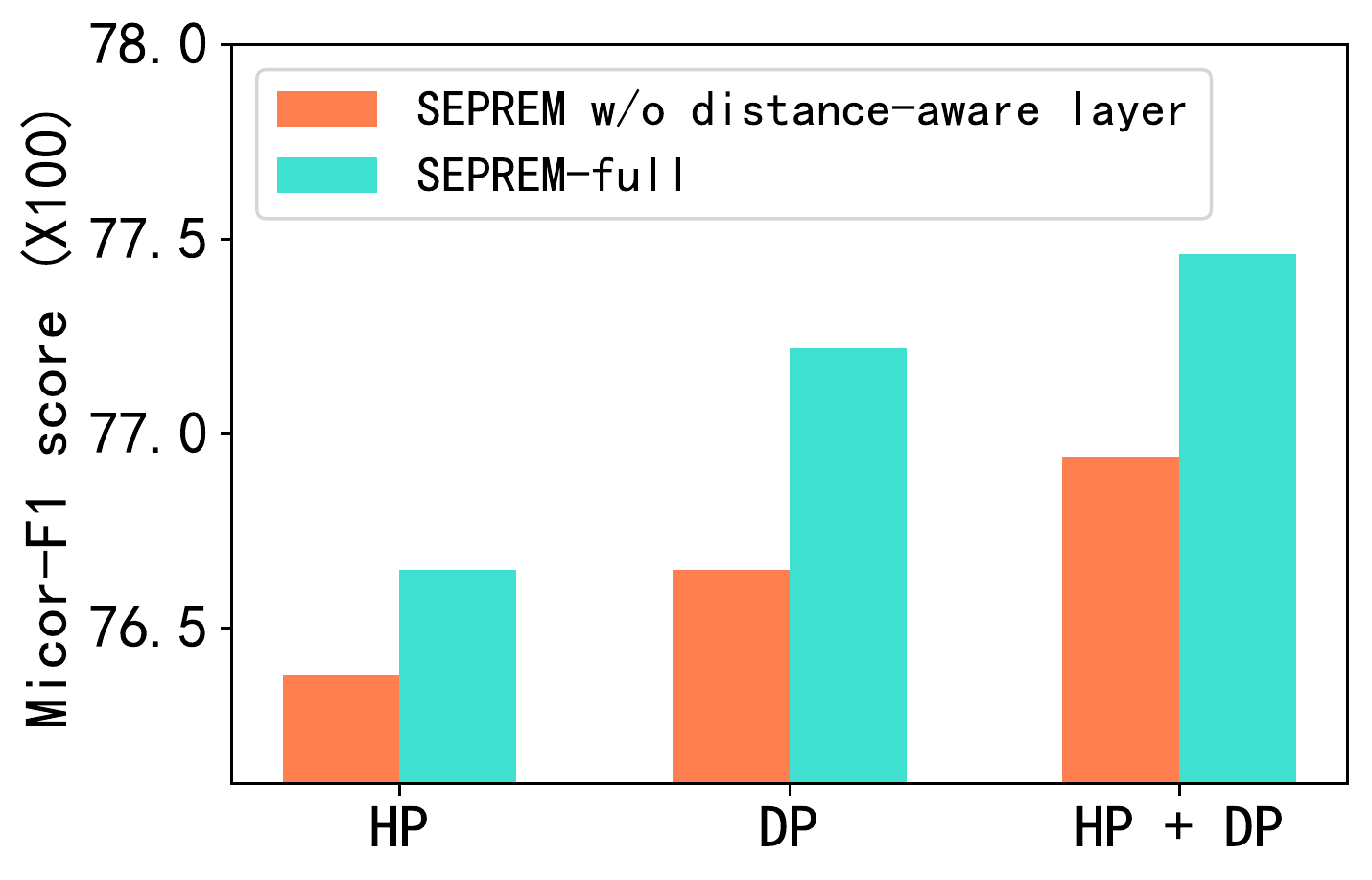}
			\end{minipage}
		}%
		\subfigure[CosmosQA]{
			\begin{minipage}[t]{0.33\linewidth}
				\centering
				\includegraphics[width=2.05in]{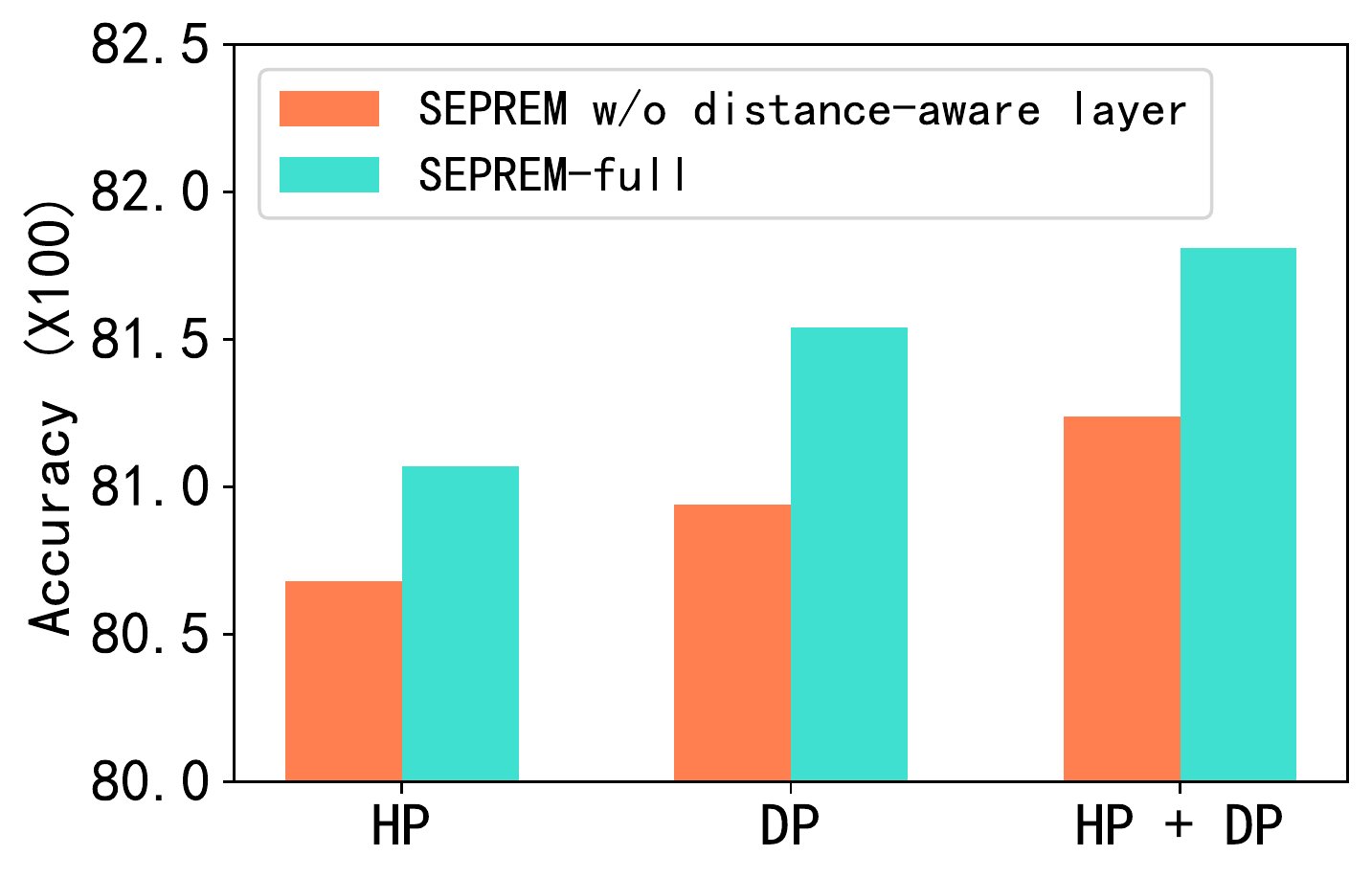}
			\end{minipage}
		}%
		\centering
		\subfigure[TACRED]{
			\begin{minipage}[t]{0.33\linewidth}
				\centering
				\includegraphics[width=2.06in]{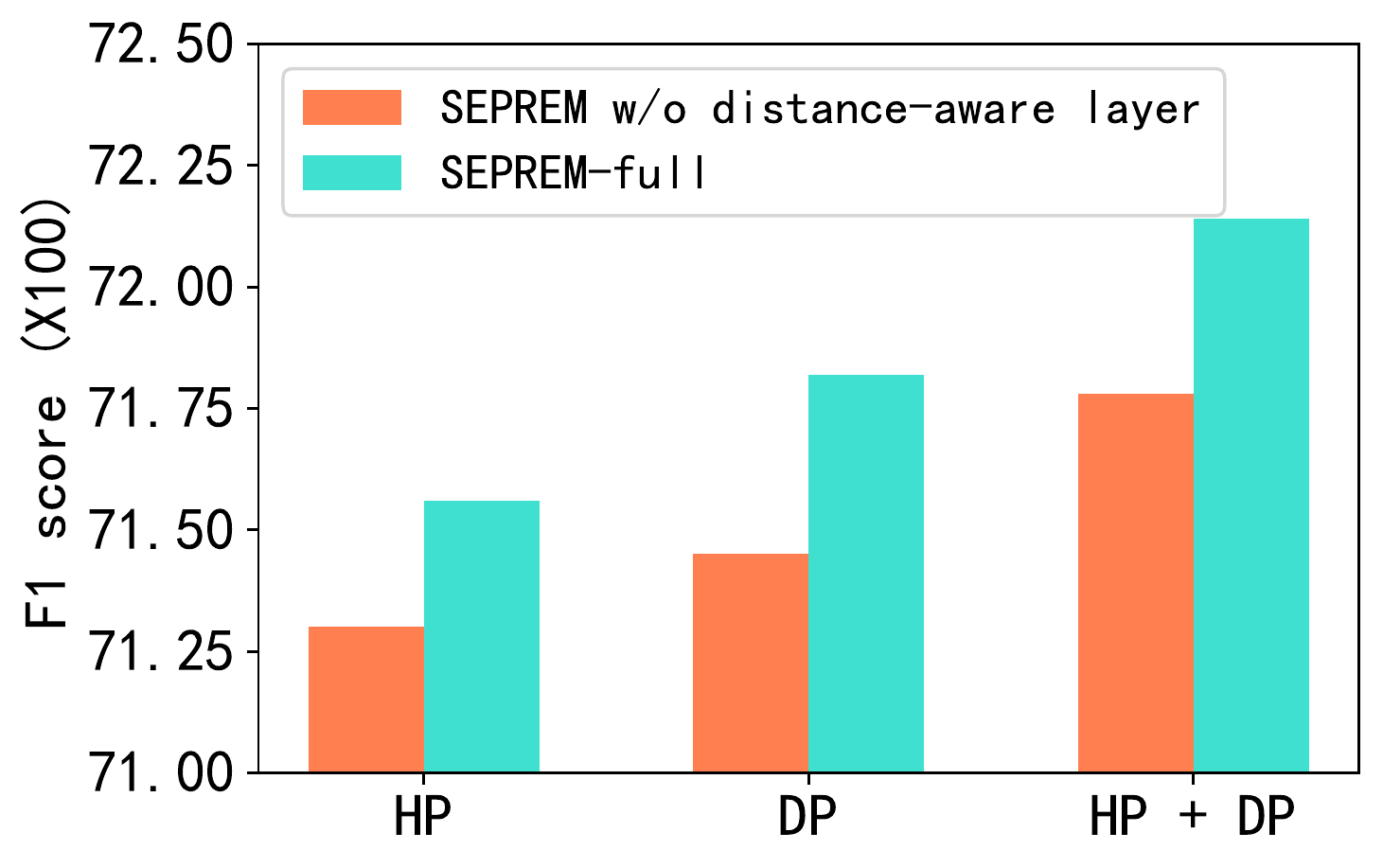}
			\end{minipage}%
		}%
		\caption{Ablation study of the SEPREM model on three different datasets over entity typing, question answering, and relation classification tasks. All the evaluation models are pre-trained on 10 million sentences.}
		\label{fig:ablation}
		\vskip -0.1in
	\end{figure*}
	
	\begin{figure*}[ht]
		\centering
		\includegraphics[scale=0.43]{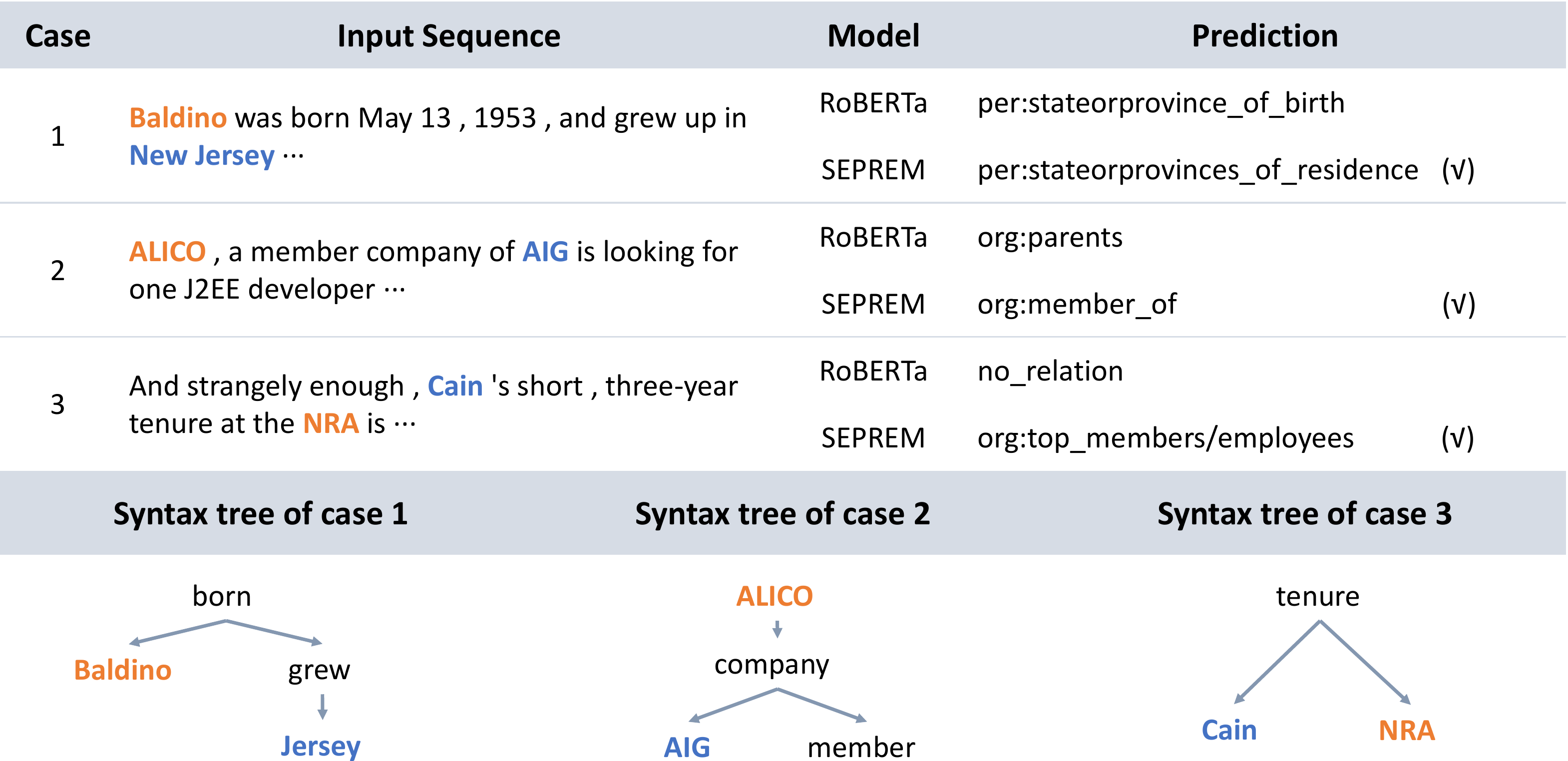}
		\caption{Case study results on the TACRED dataset of relation classification tasks. Models are required to predict the relation between tokens in 
			\bm{{\color{BurntOrange}orange}} and \bm{{\color{NavyBlue}blue}} colors. Predictions with mark \checkmark are the same with true labels.} 
		\label{fig:CT}
		\vskip -0.1in
	\end{figure*}
	
	\subsection{Relation Classification}
	A relation classification task aims to predict the relation between two given entities in a sentence. We use a large-scale relation classification dataset TACRED \citep{zhang2017tacred} for this task, and adopt Micro-precision, recall, and F1 scores as evaluation metrics. The statistics of the TACRED datasets are shown in Table \ref{table:statistic_oe_rc}. Following \citet{wang2020k}, we add special tokens ``@" and ``\#" before and after the first and second entity respectively. Then, the representations of the former token ``@" and ``\#" are concatenated to perform relation classification.
	
	\paragraph{Baselines}
	C-GCN \citep{zhang2018graph} encodes the dependency tree via graph convolutional networks for relation classification. BERT+MTB \citep{soares2019matching} trains relation representation by matching the blanks. We also include the baseline models of BERT-base \citep{zhang2019ernie}, ERNIE \citep{zhang2019ernie}, BERT-large \citep{soares2019matching}, KnowBERT \citep{peters2019knowledge}, KEPLER \citep{wang2019kepler}, RoBERTa-large, RoBERTa-large (continue training), and K-Adapter \citep{wang2020k} for a comprehensive comparison.
	
	\paragraph{Experimental Results}
	Table \ref{table:RC} shows the performances of baseline models and the proposed SEPREM on TACRED. As we can see that the proposed syntax-aware pre-training tasks and syntax-aware attention mechanism can continuously bring gains in relation classification task and SEPREM outperforms baseline models overall. This further confirms the outstanding generalization capacity of our proposed model. It can be also seen that compared with K-Adapter model, the performance gains of SEPREM model observed in the TACRED dataset are not as substantial as that in Open Entity dataset. This may be partially due to the fact that K-Adapter also injects factual knowledge into the model, which may help in identifying relationships.
	
	\subsection{Ablation Study}
	To investigate the impacts of various components in SEPREM, experiments are conducted for entity typing, question answering and relation classification tasks under the different corresponding benchmarks, ${\it i.e.}$,  Open Entity, CosmosQA, and TACRED, respectively. Note that due to the time-consuming issue of training the models on entire data, we randomly sample 10 million sentences from the whole data to build a small dataset in this ablation study.
	
	The results are illustrated in Figure \ref{fig:ablation}, in which we eliminate two syntax-aware pre-training tasks ({\it i.e.,} HP and DP) and syntax-aware attention layer to evaluate their effectiveness. It can be seen that without using the syntax-aware attention layer, immediate performance degradation is observed, indicating that leveraging syntax-aware attention layer to learn syntax-aware representation could benefit the SEPREM. Another observation is that for all three experiments, eliminating DP pre-training task leads to worse empirical results. In other words, compared with existing method ({\it i.e.}, head prediction task), the proposed dependency distance prediction task is more advantageous to various downstream tasks. This observation may be attributed to the fact that leveraging global syntactic correlation is more beneficial than considering local correlation. Moreover, significant performance gains can be obtained by simultaneously exploiting the two pre-training tasks and syntax-aware attention layer, which further confirms superiority of our pre-training architecture.

	\subsection{Case Study}
	We conduct a case study to empirically explore the effectiveness of utilizing syntax information. In the case of relation classification task, we need to predict the relationship of two tokens in a sentence.
	As the three examples shown in Figure \ref{fig:CT}, SEPREM can capture the syntax information by the dependency tree and make correct predictions. However, without utilizing syntax information, RoBERTa fails to recognize the correct relationship. To give further insight of how syntax information affects prediction, we also take case 1 for detailed analysis. The extracted dependency tree captures the close correlation of ``\textit{grew}'' and ``\textit{Jersey}'', which indicates that ``\textit{New Jersey}'' is more likely to be a residence place. These results reflects that our model can better understand the global syntax relations among tokens by utilizing dependency tree.
	
	\subsection{Analysis of Importance Score $\alpha$}
	\label{model analysis}
	Under the syntax-enhanced pre-trained framework introduced here, the contextual representation ($\bm{H}^n$) and syntax-aware representation ($\bm \hat{H}^n$) are jointly optimized to abstract semantic information from sentences. An interesting question concerns how much syntactic information should be leveraged for our pre-trained model. In this regard, we further investigate the effect of the importance score $\alpha$ on the aforementioned six downstream tasks, and the learned weights $\alpha$ after fine-tuning SEPREM model are shown in Table \ref{table:alpha}. We observe that the values of $\alpha$ are in the range of 13\% and 15\% on six downstream datasets, which indicates that those downstream tasks require syntactic information to obtain the best performance and once again confirms the effectiveness of utilizing syntax information.
	
	To have a further insight of the effect brought by importance score $\alpha$, we conduct experiments on SEPREM w/o $\alpha$, which eliminates the $\alpha$ in Equation \ref{equ:transformer} and equally integrates the syntax-aware and contextual representation, i.e., $\bm{H}^n=transformer_n(\bm{H}^{n-1}+\bm{\hat{H}}^{n-1})$. The pre-training settings of the SEPREM w/o $\alpha$ model are the same with the proposed SEPREM model. It can be seen in Table \ref{table:alpha} that, the performances drop 1\%$\sim$3\% on the six datasets when excluding the $\alpha$. This observation indicates the necessity of introducing the $\alpha$ to better integrate the syntax-aware and contextual representation.
	
		\begin{table}[!tp]
		\resizebox{\linewidth}{!}{
			\begin{tabular}{|c|c|c|c|}
				\toprule
				Datasets & Model & Performance & Values of $\alpha$ \\
				\midrule
				\multirow{2}{*}{Open Entity} & SEPREM & 79.06 & 0.1334 \\
				\cmidrule(r){2-4} & SEPREM w/o $\alpha$ & 77.13 & - \\
				\midrule
				\multirow{2}{*}{FIGER} & SEPREM & 82.05 & 0.1428 \\
				\cmidrule(r){2-4} & SEPREM w/o $\alpha$ & 79.54 & - \\
				\midrule
				\multirow{2}{*}{SearchQA} & SEPREM & 67.74 & 0.1385 \\
				\cmidrule(r){2-4} & SEPREM w/o $\alpha$ & 66.31 & - \\
				\midrule
				\multirow{2}{*}{Quasar-T} & SEPREM & 53.18 & 0.1407 \\
				\cmidrule(r){2-4} & SEPREM w/o $\alpha$ & 51.84 & - \\
				\midrule
				\multirow{2}{*}{CosmosQA} & SEPREM & 82.37 & 0.1357 \\
				\cmidrule(r){2-4} & SEPREM w/o $\alpha$ & 81.06 & - \\
				\midrule
				\multirow{2}{*}{TACRED} & SEPREM & 72.42 & 0.1407 \\
				\cmidrule(r){2-4} & SEPREM w/o $\alpha$ & 71.82 & - \\
				\bottomrule
			\end{tabular}
		}
		\caption{The model's performance and the corresponding values of importance score $\alpha$ after fine-tuning on six public benchmark datasets. Performance is under the evaluate metrics of either Mi-F1 or accuracy scores.}
		\label{table:alpha}
		\vskip -0.18in
	\end{table}
	
	\section{Conclusion}
	In this paper, we present SEPREM that leverage syntax information to enhance pre-trained models. To inject syntactic information, we introduce a syntax-aware attention layer and a newly designed pre-training task are proposed. Experimental results show that our method achieves state-of-the-art performance over six datasets. Further analysis shows that the proposed dependency distance prediction task performs better than dependency head prediction task.
	
	\section*{Acknowledgments}
	We are grateful to Yeyun Gong, Ruize Wang and Junjie Huang for fruitful comments. We are obliged to Zijing Ou and Wenxuan Li for perfecting this article. We appreciate Genifer Zhao for beautifying the figures of this article. 
	Zenan Xu and Qinliang Su are supported by the National Natural Science Foundation of China (No. 61806223, 61906217, U1811264), Key R\&D Program of Guangdong Province (No. 2018B010107005), National Natural Science Foundation of Guangdong Province (No. 2021A1515012299). Zenan Xu and Qinliang Su are also supported by Huawei MindSpore.
	
\bibliographystyle{acl_natbib}
\bibliography{anthology,acl2021}
\end{document}